\documentclass{article}
\usepackage{ijcai20}

\usepackage{times}
\usepackage{soul}
\usepackage{url}
\usepackage[hidelinks]{hyperref}
\usepackage[utf8]{inputenc}
\usepackage[small]{caption}
\usepackage{graphicx}
\usepackage{amsmath}
\usepackage{amsthm}
\usepackage{amsfonts}
\usepackage{booktabs}
\usepackage{algorithm}
\usepackage{algorithmic}
\usepackage{color,xcolor}
\usepackage{setspace}

\newcommand{\xjc}[1]{\textcolor[rgb]{0,0,0}{#1}}
\newcommand{\gjy}[1]{\textcolor[rgb]{0,0,0}{#1}}

\title{SceneEncoder: Scene-Aware Semantic Segmentation of Point Clouds \\ with A Learnable Scene Descriptor}

\author{Jiachen Xu,$^{1*}$ Jingyu Gong,$^{1}$\thanks{Equal Contribution.} Jie Zhou,$^{1}$ Xin Tan,$^{1}$ Yuan Xie,$^{2\dagger}$ Lizhuang Ma,$^{1, 2}$\thanks{Corresponding Author.}\\
\affiliations
$^1$Department of Computer Science and Engineering, Shanghai Jiao Tong University, Shanghai, China\\
$^2$School of Computer Science and Software Engineering, East China Normal University, Shanghai, China\\
\emails
\{xujiachen, gongjingyu, lord\_liang, tanxin2017\}@sjtu.edu.cn, xieyuan8589@foxmail.com, ma-lz@cs.sjtu.edu.cn
}

\begin{document}

\maketitle

\begin{abstract}
Besides local features, global information plays an essential role in semantic segmentation, while recent works usually fail to explicitly extract the meaningful global information and make full use of it. In this paper, we propose a SceneEncoder module to impose a scene-aware guidance to enhance the effect of global information. The module predicts a scene descriptor, which learns to represent the categories of objects existing in the scene and directly guides the point-level semantic segmentation through filtering out categories not belonging to this scene. Additionally, to alleviate segmentation noise in local region, we design a region similarity loss to propagate distinguishing features to their own neighboring points with the same label, leading to the enhancement of the distinguishing ability of point-wise features. We integrate our methods into several prevailing networks and conduct extensive experiments on benchmark datasets ScanNet and ShapeNet. Results show that our methods greatly improve the performance of baselines and achieve state-of-the-art performance.

\end{abstract}

\section{Introduction}
\label{sec:intro}

As a basic task in 3D vision, semantic segmentation of point clouds has drawn more and more attention. Due to the irregularity and disorder of point clouds, many previous works convert point clouds into the grid representation through voxelization or projection to leverage the effectiveness of grid convolution~\cite{zhou2018voxelnet,su2015multi}. These methods inevitably destroy the original geometric information of point clouds. Therefore, PointNet~\cite{qi2017pointnet} directly processes the raw point clouds and extracts features with shared Multi-Layer Perceptrons (MLPs). However, because of the unusual properties of point clouds and the complexity of scene segmentation, semantic segmentation of point clouds remains a challenging issue.

\begin{figure}[t]
    \centering
    \includegraphics[width=0.9\linewidth]{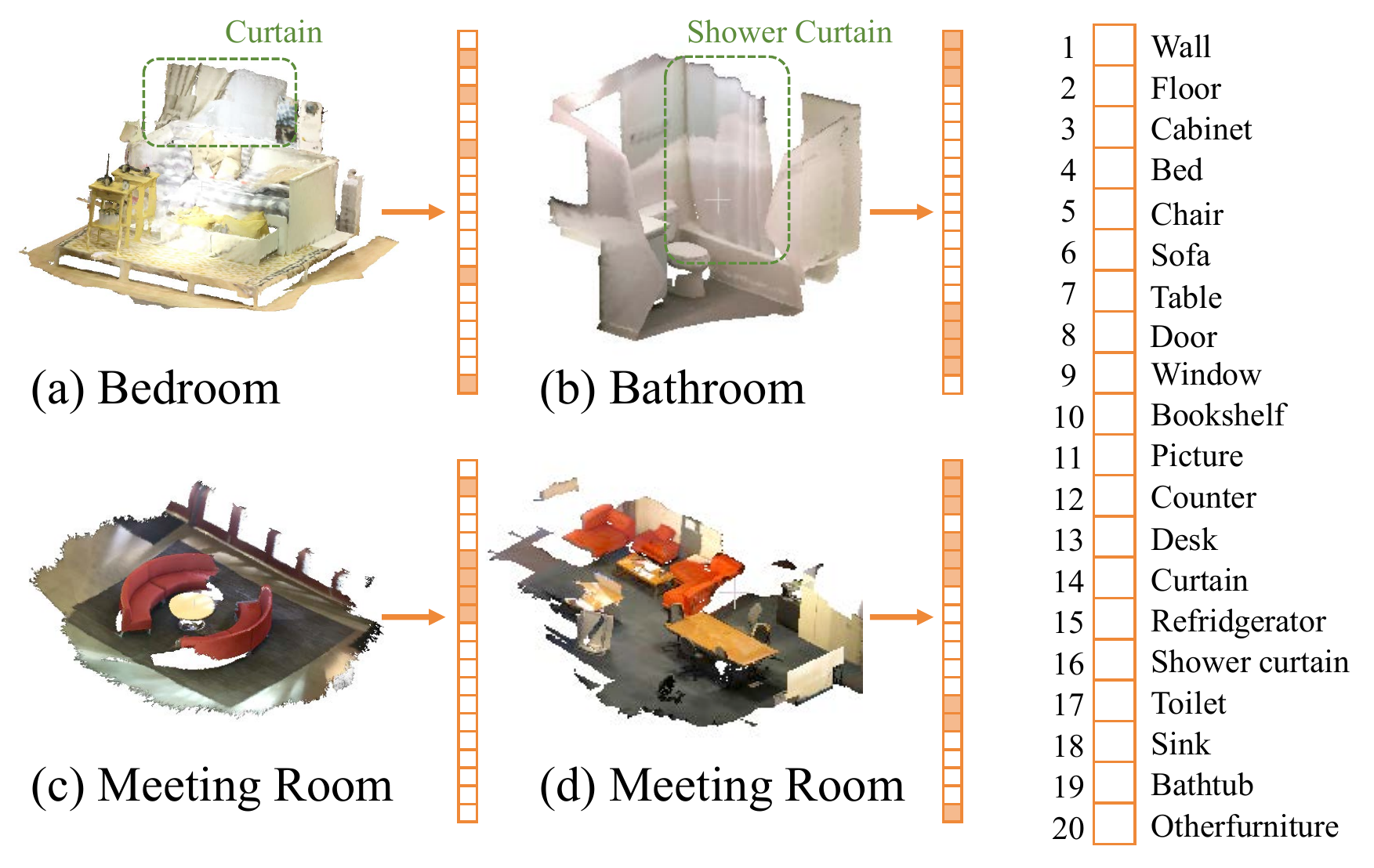}
    \vspace{-2.5mm}
    \caption{Illustration of scene descriptors for different scenes. The target scene descriptor is a binary vector for each scene. The $i$-th element of scene descriptor is 1 if the $i$-th category exists in the scene and 0 if not. Make an analogy with grids. The colored grids represent 1 and the empty grids represent 0.}
    \vspace{-2.5mm}
    \label{fig:fig1}
\end{figure}

{\color{black}Global information, which is commonly extracted by the encoder network, usually contains the essential knowledge that directly summarizes the information of the whole scene}, thus should play a more significant role in the semantic segmentation. For human beings, the prior knowledge towards a scene could directly influence the semantic comprehension. {\color{black}For instance,} as in Figure \ref{fig:fig1}, {\color{black}when being in a bathroom,} it is easy to distinguish the ``shower curtain" category from the ``curtain" category even they look similar. \emph{Therefore, the global information of {\color{black}point clouds} can play the role of prior knowledge for semantic segmentation.} 

Many previous works utilize the U-Net architecture to extract global features with growing receptive field{\color{black},} and then mix these global features with local features by concatenation hierarchically \cite{wu2019pointconv,graham20183d}. 
{\color{black}However, such a kind of mixture not only attenuates the guidance of global information, but also degrades the representation capability of local features.}
SpiderCNN~\cite{xu2018spidercnn} adopts a more direct way to take the global label of {\color{black}a point cloud} as {\color{black}a part of elements in feature vectors} to enhance the effect of the global information. 
However, {\color{black}it is based on an impractical assumption that the global label is always available, especially in testing phase.}

{\color{black}Therefore, we propose a SceneEncoder module to predict a multi-hot scene descriptor for every point cloud, a novel representation of global label rather than the manually annotated one-hot scene label, to perform a scene-level guidance.} 
An ideal scene descriptor is designed to be a {\color{black}binary} vector {\color{black}with each element representing the existence of the corresponding category of object}.
As shown in Figure \ref{fig:fig1}(a), the scene descriptor of a bedroom, consisting of floor, bed, table, curtain and some other furniture, can be represented by a 5-hot vector. 

Different from concatenating global features to local features, 
we utilize the scene descriptor as a {\color{black}mask/attention} to aid the point-level semantic segmentation by filtering out {\color{black}irrelevant object categories that are impossible to exist in the current scene.} 
For example, there is a curtain and a shower curtain in Figure \ref{fig:fig1}(a) and (b) respectively, which are difficult to distinguish {\color{black}due to the similar appearance.} However, {\color{black}our descriptors} can directly help to exclude the classification option of the shower curtain in bedroom scenes and exclude the curtain in the bathroom. Besides, compared with a one-hot scene label, our well-designed multi-hot scene descriptor is also able to subdivide each category of scenes. As shown in Figure \ref{fig:fig1}(c) and (d), meeting rooms which are designed for different purposes can be further represented by different scene descriptors. 
To train the SceneEncoder module, we dynamically generate the ground truth of scene descriptors by checking which categories of objects exist in the input training scene point clouds based on labels of all points.

Since the segmentation task is a point-level task, global information is not enough for point-level classification. Therefore, an increasing number of works~\cite{zhao2019pointweb,wu2019pointconv} focus on exploiting the local context which is important for recognizing fine-grained patterns and generalizing to complex scenes. {\color{black}A problem in the prevailing methods is that the extracted feature usually involves the features of different categories in the local region.}
{\color{black}For example, the point clouds belonging to a chair might be nearby a table, such that the extracted local feature would incorporate features of both the chair and table. This would make} point features less distinguishing because these features encode more than one class information, and it also results in poor object contours in the segmentation task.
To tackle this problem, we design a novel loss, {\color{black}namely region similarity loss}, to propagate distinguishing point features to ambiguous features in local regions. That is, we only propagate each distinguishing feature to its adjacent points ({\it {\color{black}i.e.,}} points in its neighborhood) with same labels since the points in the same local region but with different categories would differ on their features. 
{\color{black}Overall, the major contributions can be summarized as follows:}






\begin{itemize}
    \item {\color{black}We design a SceneEncoder module with a scene descriptor to provide a scene-level guidance, leading to an effective collaboration of the global information and {\color{black}local features} for semantic segmentation.} 
    \item We propose a novel region similarity loss to propagate distinguishing point features to ambiguous features in local regions {\color{black}so as to alleviate the segmentation noise effectively.} 
    \item {\color{black}We integrate our SceneEncoder module into two prevailing networks by involving the region similarity loss. The experimental results on two benchmark datasets demonstrate that our method outperforms many state-of-the-art competitors.}
\end{itemize}

\begin{figure*}
    \centering
    \includegraphics[width=\linewidth]{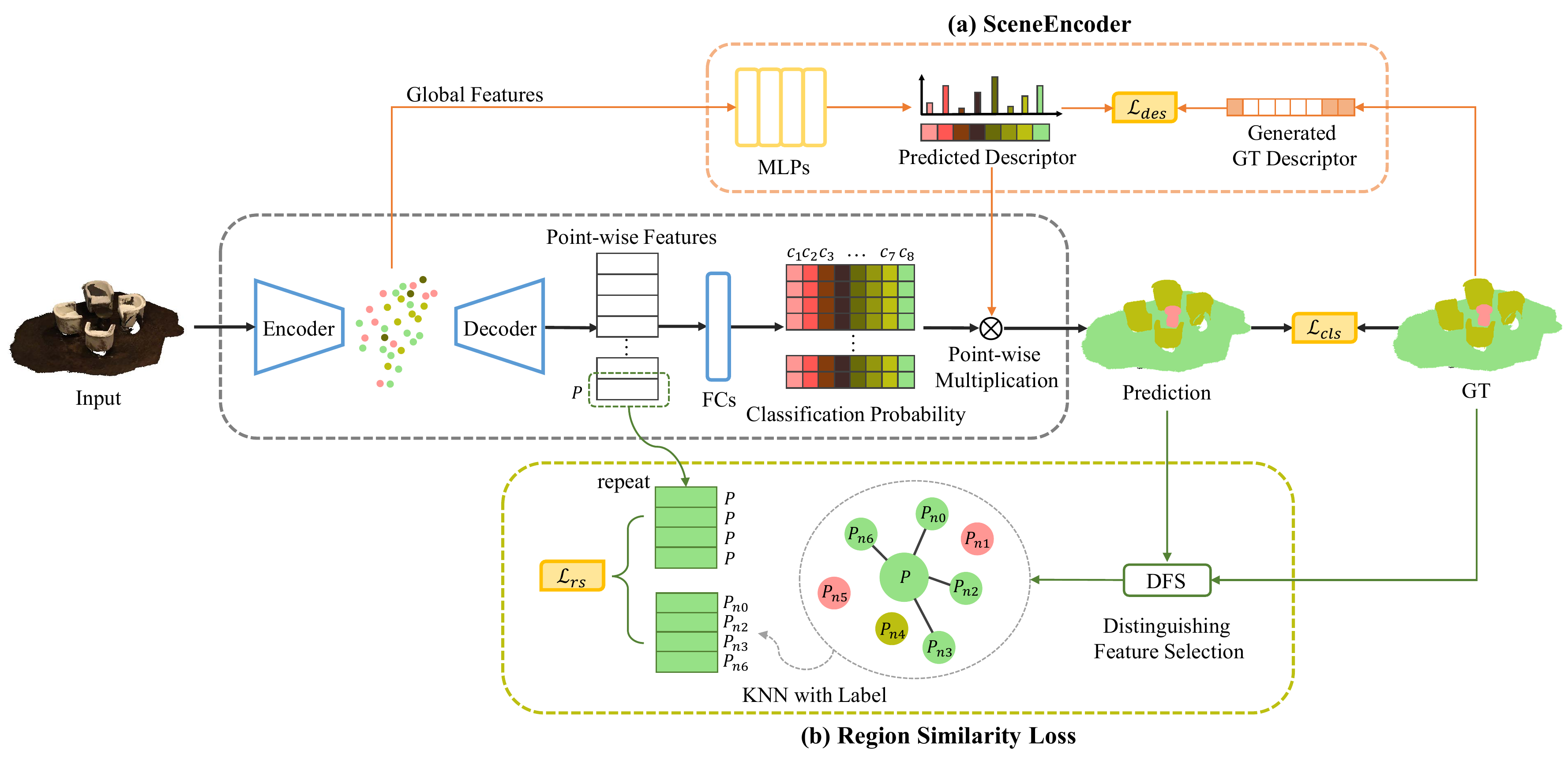}
    \vspace{-5mm}
    \caption{Overall architecture of our network. (a) represents the SceneEncoder module in which the predicted descriptor is directly regularized by a well-designed scene descriptor and aids the final segmentation. (b) stands for the region similarity loss in which distinguishing features help to refine other ambiguous features.}
    \vspace{-5mm}
    \label{fig:framework}
\end{figure*}

\section{Related Work}
\label{sec:related}

\paragraph{Segmentation on point clouds.} Impressively, PointNet~\cite{qi2017pointnet} processed the raw point clouds directly and extracted point features with MLPs. {\color{black}Built upon it, PointNet++~\cite{qi2017pointnet++} {\color{black}designed} a hierarchical framework to exploit local context with growing receptive fields. However, the key operation to aggregate features in both methods} is max-pooling, which leads to great loss of context information. 

To utilize the local context efficiently, 
PointWeb~\cite{zhao2019pointweb} densely connected each pair of points in the local region and adjusted each point feature adaptively. An independent edge branch was introduced in ~\cite{jiang2019hierarchical} to exploit the relation between neighboring points at different scale and interwove with the point branch to provide more context information. PointConv~\cite{wu2019pointconv} focused on the distribution of points in the local region and utilized density information. Furthermore, an octree guided selection strategy was proposed in ~\cite{lei2019octree} to guide the convolution process {\color{black} of point clouds} in a more reasonable way. Compared with these methods {\color{black} which utilize the scene information by fusing local and global features together}, we directly {\color{black}predict} a scene descriptor from global features to {\color{black}guide} the point-level segmentation so as to enhance the effect of the global information. 
\paragraph{Segmentation Refinement.} Additionally, {\color{black}noises and poor contours are usually caused by ambiguous features because they are mixed with the features of different categories in the local region.}
SEGCloud~\cite{tchapmi2017segcloud} {\color{black}proposed} to combine networks with fine-grained representation of point clouds using 3D Conditional Random Fields (CRF). GAC~\cite{wang2019graph} determined how each point feature contribute to the extracted feature by the similarity between these two features to improve the distinguishability of features. {\color{black}By contrast, we propose a more general loss to refine the results of segmentation by propagating distinguishing features in the local region. This strategy could be adopted in the training process of most networks in semantic segmentation task.}
\paragraph{Loss Function.} In ~\cite{de2017semantic}, a widely used loss was proposed to minimize the difference between point features of the same instance object. Similarly, Pairwise Similarity Loss~\cite{engelmann2018know} maximized the similarity {\color{black}among} point features of the same semantic object. {\color{black}It is noteworthy that points which are far apart should have different features even if they belong to the same category.} Therefore, our loss just focus on forcing adjacent point features of the same category to be similar.

\section{Method}
\label{sec:method}
First we introduce the overall architecture in Sec. \ref{subsec:overview}. {\color{black}Then, the SceneEncoder module and the scene descriptor will be described in details in Sec. \ref{subsec:descriptor}. In Sec. \ref{subsec:rsl}, the novel region similarity loss would be depicted. Finally, we summarize different losses used in the training process of the whole network in Sec. \ref{subsec:loss}.}


\subsection{Overview}
\label{subsec:overview}
Figure \ref{fig:framework} illustrates our overall architecture. {\color{black}We encode the raw point cloud into few global features, which will be taken as the input into the proposed SceneEncoder module (Figure \ref{fig:framework}(a)). The output of this module, {\it i.e.,} the predicted scene descriptor, would be used to point-wise multiply with classification probability vector of each point, so as to exclude the impossible category options and get final prediction.} Furthermore, during the training process, {\color{black}we select $M$ distinguishing points that are correctly classified and their corresponding features from the feature map. Then, we define a region similarity loss (Figure \ref{fig:framework}(b)) to {\color{black}propagate distinguishing features by improving the feature similarity between each distinguishing point and its neighboring points with the same label.}}

\subsection{SceneEncoder Module}
\label{subsec:descriptor}
The global information {\color{black}can give a top-down guidance to point-level classifier. To make full use of the global information as the scene prior knowledge,} 
we propose a SceneEncoder module with a well-designed multi-hot scene descriptor to enhance the effect of global information in the semantic comprehension. {\color{black}This descriptor is able to substitute the global scene label, and acts as a mask/attention to eliminate impossible results when classifying each point.}
\paragraph{Multi-hot Scene Descriptor.} {\color{black}As} the recognition of a scene can directly influence the result of point-level semantic segmentation, we propose to use a scene descriptor to represent the global scene information. {\color{black}For scene semantic segmentation task with $n$ object categories, we design a multi-hot vector of length $n$ with each element representing the probability of existence of the corresponding category in the point cloud. Specifically, let $\widetilde{\mathbf{g}}$ denote the predicted descriptor,} the $i$-th element of this descriptor $\widetilde{\mathbf{g}}_i$ will be 0 if the $i$-th category does not exist in this point cloud. Then, in semantic segmentation, we could apply softmax on the output to obtain the {\color{black}classification probability map $\widetilde{P} \in \mathbb{R}^{N\times n}$, where $N$ denotes the number of points, $n$ is the number of classes, and $\widetilde{P}(i,j)$ represents the probability that point $i$ belongs to class $j$. To make the scene descriptor directly guide the semantic segmentation, we directly multiply the probability map $\widetilde{P}$ with the predicted descriptor $\widetilde{\mathbf{g}}$ to achieve the refined probability map $\widetilde{P}_{ref}$ as follow.}
\begin{equation}
\label{eq:multiply}
\widetilde{P}_{ref}(i,j) = \frac{\widetilde{\mathbf{g}}_j \cdot \widetilde{P}_{(i,j)}}{\sum_{j=1}^{n}\widetilde{\mathbf{g}}_j \cdot \widetilde{P}_{(i,j)}}
\end{equation}
where {\color{black}$\widetilde{P}_{ref}(i, j)$} is the final predicted probability that point $i$ belongs to class $j$, and $\widetilde{\mathbf{g}}_j$ indicates the probability of existence of the class $j$ in the whole scene.

{\color{black}In this way, global information plays an important and influential role of prior knowledge in semantic segmentation by filtering out impossible results. Obviously, two scenes with the same descriptor are likely to be the same type of scene, and this makes our descriptor be able to substitute the scene label. Furthermore, compared with the manually annotated label, our scene descriptor could subdivide each scene according to {\color{black}the object composition} and provide a more concrete global information.}

\paragraph{Supervision for Scene Descriptor.} Compared with the SpiderCNN~\cite{xu2018spidercnn} that {\color{black}requires} the ground truth of global label during both training and testing, we generate the scene descriptor {\color{black}through the SceneEncoder module. To effectively train the SceneEcoder module, we regularize the predicted scene descriptor $\widetilde{\mathbf{g}}$.} Instead of labeling the ground truth of each scene descriptor $\mathbf{g}$ manually, {\color{black}we generate it on-the-fly from the labels of all points in the point cloud during the training process.} Note that, the 
ground truth of the scene descriptor is a binary vector of length $n$, and each element represents the existence of corresponding category in this scene. Therefore, through checking the labels of all the points, we can {\color{black}confirm} which classes appear in this point cloud easily. Then, the regularization is as follow:
\begin{equation}
\label{eq:des}
    \mathcal{L}_{des} = -\sum_{j=1}^{n}\mathbf{g}_j log(\widetilde{\mathbf{g}_j}){\color{black}.} 
\end{equation}

In the training phase, the SceneEncoder module could learn to represent the type of each scene {\color{black}by encoding the global information into the scene descriptor. As for the validation and testing process,} the scene descriptor could be predicted by the well-trained SceneEncoder module and directly aid point-level prediction through Eq. (\ref{eq:multiply}).

\subsection{Region Similarity Loss}
\label{subsec:rsl}
Compared with global information, local details could help to generalize to complex scenes. However, {\color{black}during} feature aggregation process, previous works ignore the difference among points with different labels in each local context. {\color{black}Such a kind of label inconsistency will lead} to the lack of distinguishing ability of point features. {\color{black}Consequently}, there are usually poor contours and noisy regions in the segmentation results. 
Therefore, we propose a novel region-based loss to propagate distinguishing features in local regions, thus those nondistinguishing features of neighbors with the same label can be refined. 

\paragraph{Distinguishing Feature Selection.} 

{\color{black}Compared with features achieved from intermediate layers, point features of the last layer encode more local and global information, and directly affect the segmentation results.} Hence, our proposed loss directly refines point features of the last feature abstraction layer. 

As shown in Figure \ref{fig:framework}, to select distinguishing point features, we first {\color{black}choose} a set of correctly classified points $\mathcal{P}$. Then, among these points, we analyze two strategies to select a fixed number $M$ of points. {\color{black}The first strategy is randomly selection, while the second one is to pick points with the top $M$ classification confidence from $\mathcal{P}$.} However, the first strategy is too arbitrary so that some correctly classified points with low confidence may also be selected, {\color{black}leading to incorporate indistinguishable features.}

{\color{black}By contrast, the second strategy is more reasonable, which will be confirmed in Sec. \ref{subsec:ablation}. In practice, if the amount of correctly classified points is less than the given number $M$, which usually happens in the beginning of training process, the point feature with the highest classification confidence is sampled repeatedly.} 

\paragraph{Feature Adjustment.} Based on {\color{black}the selected distinguishing point features, we adjust other features to boost their distinguishing ability.} First, since points of different categories should have different features, we only propagate distinguishing features to points of the same category. Then, we achieve the feature propagation by reducing the difference between each distinguishing feature and features of points in its local neighborhood instead of the whole scene. Because in most cases, features of points that are far away from each other are likely to be dissimilar even if they belong to the same category. If we force to reduce the difference between two features but their corresponding points are far apart, it would hinder the feature abstraction layers from learning the correct point features, {\color{black}making the deep network hard to train.}

Therefore, as shown in Figure \ref{fig:framework}(b), we {\color{black}utilize} a KNN with label strategy to select neighboring points of same category, and only propagate each distinguishing feature in corresponding neighborhood. {\color{black}Concretely}, for each point $p_i$ with distinguishing feature, its $k$ nearest neighbors $p_{n_{i1}}, p_{n_{i2}}, ..., p_{n_{ik}}$ with the same label are adjusted. To minimize the difference between each neighboring point feature and the center distinguishing point feature, {\color{black}we use the cosine similarity to define the region similarity loss as follows:}

\begin{equation}
\label{eq:rs}
\begin{split}
    \mathcal{L}_{rs}&=-\frac{1}{M}\sum_{i=1}^M\sum_{j=1}^k cos\_sim({f_{p_i}}, f_{p_{n_{ij}}}) \\ &=-\frac{1}{M}\sum_{i=1}^M\sum_{j=1}^k\frac{f_{p_i}\cdot f_{p_{n_{ij}}}}{max(\Arrowvert f_{p_i}\Arrowvert_2\cdot\Arrowvert f_{p_{n_{ij}}}\Arrowvert_2,\epsilon)}
\end{split}
\end{equation}
where $M$ is the number of distinguishing point features and $k$ is the number of selected neighboring points. ${f_{p_i}}$ and $f_{p_{n_{ij}}}$ are features of center point and neighboring point, respectively.

In addition to cosine similarity, {\color{black}Euclidean distance can also be used as a distance measurement. However, it is sensitive to the value of each dimension of each point feature, reducing generalization ability of the loss. On the contrary,} cosine similarity could make our proposed loss more robust to point clouds in different scales.

\subsection{Total Loss}
\label{subsec:loss}
{\color{black}In summary,} the total loss function consists of three parts: $\mathcal{L}_{cls}$, $\mathcal{L}_{des}$, and $\mathcal{L}_{rs}$. $\mathcal{L}_{cls}$ is the cross entropy loss applied to constrain the point-level predictions. $\mathcal{L}_{des}$ {\color{black}is defined in} Eq. (\ref{eq:des}) to improve the performance of SceneEncoder module. Furthermore, $\mathcal{L}_{rs}$ is the proposed region similarity loss to {\color{black}boost the distinguishing ability of point features as in} Eq. (\ref{eq:rs}). Therefore, the total loss function is as follows:
\begin{equation}
    \mathcal{L}=\lambda_1\mathcal{L}_{cls}+\lambda_2\mathcal{L}_{des}+\lambda_3\mathcal{L}_{rs},
\end{equation}
where $\lambda_1$, $\lambda_2$ and $\lambda_3$ adjust ratio of the three losses.

In the training process, we inhibit the gradient back propagation from $\mathcal{L}_{cls}$ to $\widetilde{D}$, that is to prevent the scene descriptor from being regularized by $\mathcal{L}_{cls}$. Because, $\mathcal{L}_{cls}$ is a point-level classification loss while our scene descriptor is a global descriptor, which would lower the training performance. \gjy{In our experiments, $\lambda_1$ and $\lambda_2$ are set the same, but we dynamically change $\lambda_3$ during training.} In the beginning of training process the selected features are not distinguishing enough, so the impact to other features should be weakened. \xjc{Therefore, we assign a small weight to $\lambda_3$ in the beginning and increase the weight gradually until it is the same as $\lambda_1$ and $\lambda_2$.} In the testing process, we do not generate the ground truth of scene descriptor any more.

\section{Experiments}
Experiments consist of three parts. First, we demonstrate the effectiveness of our method on semantic segmentation task in Sec \ref{subsec:sem-seg}. Second, we generalize \gjy{our} method to part segmentation and prove its effectiveness on part segmentation in Sec \ref{subsec:part-seg}. Finally, we conduct ablation study to demonstrate the effectiveness of the proposed SceneEncoder module and region similarity loss in Sec \ref{subsec:ablation}. 
\paragraph{Metric.} Like previous works, we take the intersection-over-union (IoU) as our metric, {\color{black}{\it i.e.,}} mean IoU over categories for semantic segmentation task, class average IoU and instance average IoU for part segmentation task.

\subsection{Scene Semantic Segmentation}\label{subsec:sem-seg}

\begin{table}
\centering
\begin{tabular}{lccc}  
\toprule
Method  & xyz & rgb & mIoU \\
\midrule
PointNet++~\cite{qi2017pointnet} & \checkmark & &33.9 \\
PointCNN~\cite{li2018pointcnn} & \checkmark & \checkmark &45.8 \\
3DMV~\cite{dai20183dmv} & \checkmark & \checkmark &48.4 \\ 
PointConv~\cite{wu2019pointconv} & \checkmark &  &55.6\\
TextureNet~\cite{huang2019texturenet} &\checkmark &\checkmark & 56.6 \\
HPEIN~\cite{jiang2019hierarchical} & \checkmark & \checkmark &61.8\\
\midrule
Ours &\checkmark & &\textbf{62.8}\\
\bottomrule
\end{tabular}
\caption{Semantic segmentation results on ScanNet v2.}
\label{tab:semantic}
\end{table}

\begin{figure}
    \centering
    \includegraphics[width=\linewidth]{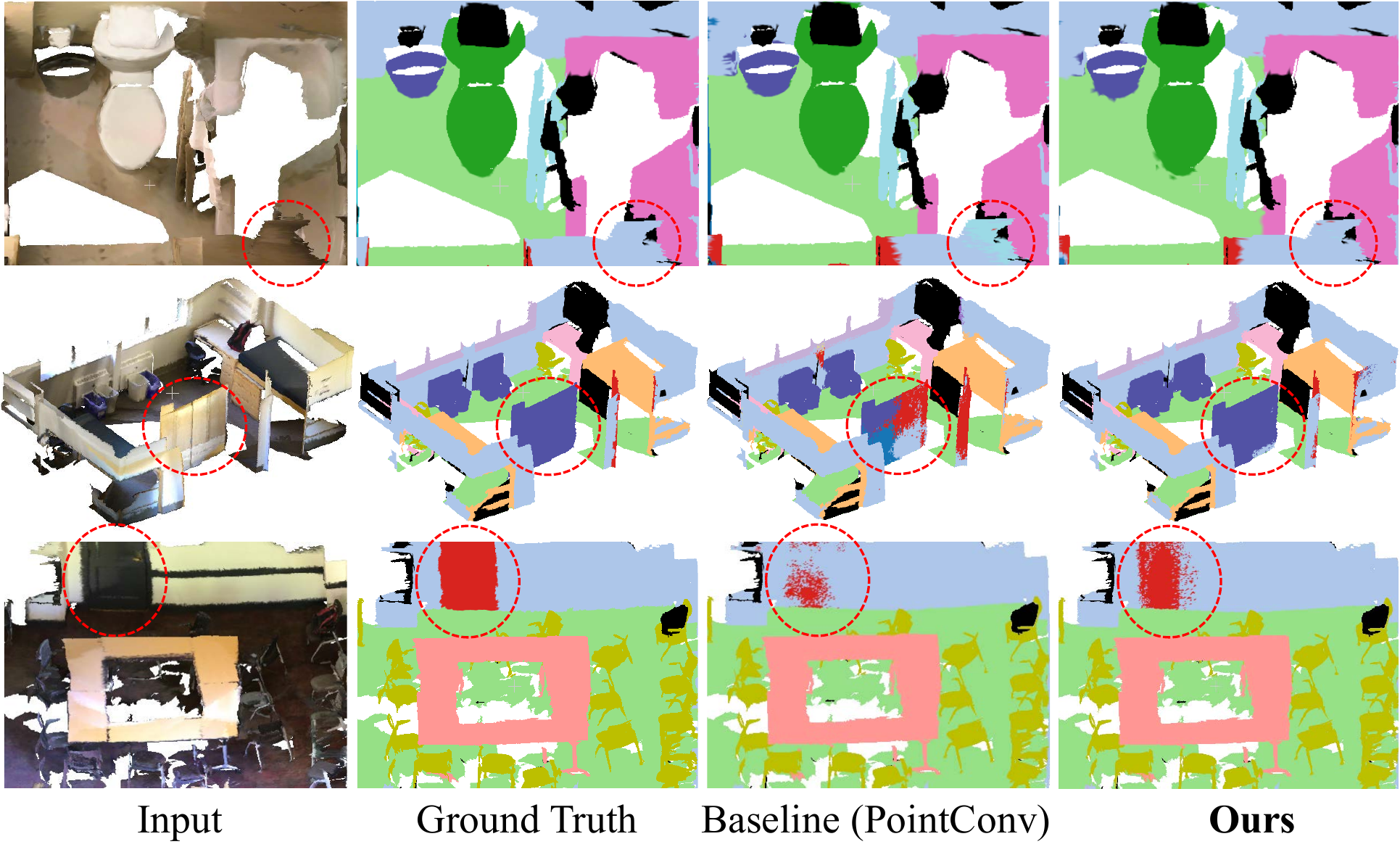}
    \vspace{-7mm}
    \caption{\gjy{Visualization results for semantic segmentation on ScanNet. The images from left to right are the original scenes, the segmentation ground truth, predictions given separately by PointConv and the proposed methods.}\iffalse Visualization of semantic segmentation results on ScanNet.\fi}
    \vspace{-3mm}
    \label{fig:scene_figure}
\end{figure}

\paragraph{Dataset.} We evaluate the performance {\color{black}of} scene semantic segmentation task on ScanNet v2~\cite{dai2017scannet}. ScanNet v2 consists of 1,201 scanned scenes for training and 312 scenes for validation. Another 100 scenes are provided as the testing dataset. {\color{black}Different from most} previous works using both geometry and color information, we just take the geometry information ({\color{black}{\it i.e.,}} xyz) as the input.

\paragraph{Implementation details.} We insert our SceneEncoder module into PointConv~\cite{wu2019pointconv}, {\color{black}in which} \gjy{point features of the last layer are regularized by our region similarity loss during training process. }
Following~\cite{wu2019pointconv}, training samples are generated {\color{black}by randomly sampling} 3m$\times$1.5m$\times$1.5m cubes from rooms, and then we {\color{black}test over} the entire scan. {\color{black}The network is trained by using Adam optimizer with batch size 8 on a single GTX 1080Ti GPU.}

\paragraph{Results.} We report mean IoU (mIoU) over categories in Table \ref{tab:semantic}, {\color{black} where a clear improvement over backbone (PointConv) can be observed, {\it i.e.,} $7.2\%$ in mIoU, and our method achieves a state-of-the-art performance in ScanNet benchmark.} As aforementioned, we only take the location information as our input and it also performs better than lots of methods that use extra color information. Figure \ref{fig:scene_figure} visualizes the scene semantic segmentation results of \gjy{ PointConv~\cite{wu2019pointconv} and our methods.}

\subsection{Part Segmentation}\label{subsec:part-seg}

\begin{table}[t]
\centering
\begin{tabular}{lcc}  
\toprule
Method  & mcIoU & mIoU \\
\midrule
PointNet++~\cite{qi2017pointnet++} & 81.9 & 85.1 \\ 
SO-Net~\cite{li2018so} & 81.0 & 84.9 \\ 
PCNN by Ext~\cite{atzmon2018point} & 81.8 & 85.1 \\
SpiderCNN~\cite{xu2018spidercnn} & 82.4 & 85.3 \\
ShellNet~\cite{zhang2019shellnet} & 82.8 & - \\
PointConv~\cite{wu2019pointconv} & 82.8 & \textbf{85.7} \\
\midrule
Ours & \textbf{83.4} & \textbf{85.7}\\
\bottomrule
\end{tabular}
\caption{Part segmentation results on ShapeNet.}
\vspace{-3mm}
\label{tab:part}
\end{table}

\begin{table*}
\begin{center}\begin{spacing}{1.19}\resizebox{\textwidth}{9.5mm}{
\begin{tabular}{lcccccccccccccccccc}  
\toprule
Method & mcIoU & mIoU & aero & bag & cap & car & chair & eph. & guitar & knife & lamp & laptop & motor & mug & pistol & rocket & skate. & table \\ 
\midrule
SpiderCNN & 81.8 & 84.8 & 82.5 & 77.2 & 85.7 & 76.9 & 90.4 & 77.0 & 90.9 & \textbf{87.6} & 81.7 & \textbf{95.8} & 68.3 & 94.6 & 81.2 & 59.4 & 76.6 & 82.8 \\
SpiderCNN + category & 82.4 & 85.3 & 83.5 & 81.0 & 87.2 & 77.5 & \textbf{90.7} & 76.8 & \textbf{91.1} & 87.3 & 83.3 & \textbf{95.8} & 70.2 & 93.5 & \textbf{82.7} & 59.7 & 75.8 & 82.8 \\
SpiderCNN + SceneEncoder & 82.4 & 85.3 & \textbf{83.9} & 80.5 & 83.9 & 78.6 & 90.6 & \textbf{80.5} & 90.7 & 87.2 & \textbf{83.8} & 95.7 & 69.7 & 94.4 & 81.6 & 58.4 & 76.5 & 82.6\\
SpiderCNN + category + SceneEncoder & \textbf{83.1} & \textbf{85.4} & \textbf{83.9} & \textbf{83.3} & \textbf{87.7} & \textbf{78.7} & 90.4 & 79.9 & \textbf{91.1} & 87.2 & 83.0 & 95.7 & \textbf{71.3} & \textbf{94.7} & 81.3 & \textbf{61.7} & \textbf{77.3} & \textbf{83.1} \\
\bottomrule
\end{tabular}}\end{spacing}\end{center}
\vspace{-7mm}
\caption{Compare the effectiveness of ground truth of category label with SceneEncoder module based on SpiderCNN. The top line removes the ground truth of category label used in SpiderCNN. Then, category label and SceneEncoder module are added into SpiderCNN separately. The bottom line shows the result where both ground truth of category label and SceneEncoder module are used in SpiderCNN.}
\vspace{-3mm}
\label{tab:spidercnn}
\end{table*}

\begin{figure}[t]
    \centering
    \vspace{-3mm}
    \includegraphics[width=1\linewidth]{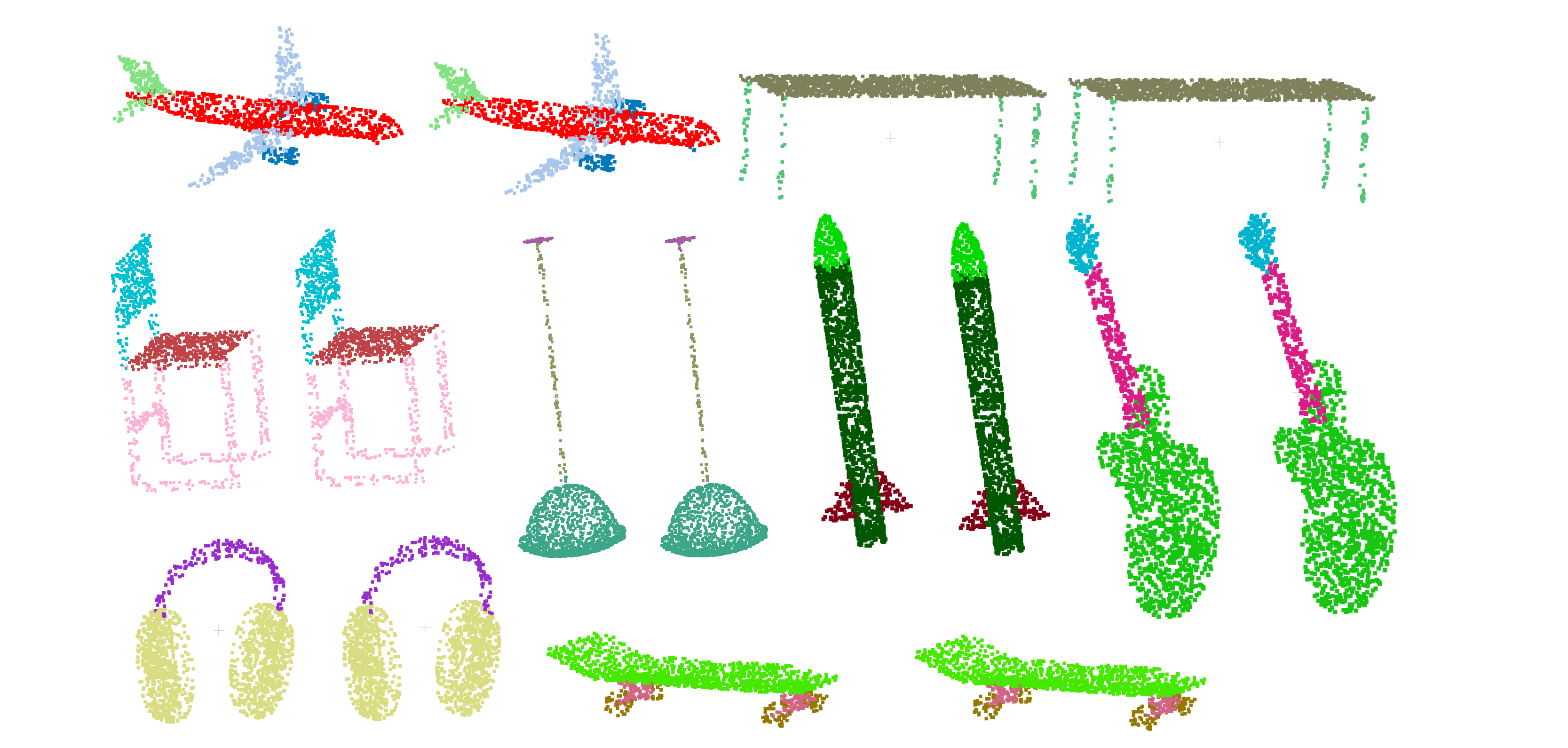}
    \vspace{-5mm}
    \caption{Visualization of part segmentation on ShapeNet. For each pair of objects, the left one is the ground truth while the right one is the predicted result.}
    \vspace{-2mm}
    \label{fig:part_figure}
\end{figure}

Similarly, our methods can also improve the performance on part segmentation. {\color{black}In part segmentation task that purposes to segment the functional part of each object, we can treat each object as the entire scene point cloud, and the parts of each object can be considered as the different objects in the scene.} Therefore, to {\color{black}confirm} the generalization ability, we evaluate our method on part segmentation.
\paragraph{Dataset.} We evaluate the performance on ShapeNet part data~\cite{chang2015shapenet}. ShapeNet part data consists of 16,881 point clouds from 16 categories and totally contains 50 parts, {\color{black}with xyz and norm vectors being taken as the input.}
\paragraph{Implementation details.} We still choose PointConv as the baseline and integrate our method into it. We directly take the raw point cloud as input without preprocessing. {\color{black}The Adam optimizer is employed to train the model with our region similarity loss on a single GTX 1080Ti GPU.}
\paragraph{Results.} We report class average IoU (mcIoU) and instance average IoU (mIoU) for part segmentation in Table \ref{tab:part}. {\color{black}The $0.6\%$ increment over PointConv in term of mcIoU shows the great potential of the proposed method in part segmentation. Overall,} the improvements on both scene semantic segmentation task and part segmentation task show the generality of our method. However, even {\color{black}obvious improvements have been shown in mcIoU, there is little increase in mIoU, seeing the rightmost column in Table \ref{tab:part}. {\color{black}The reason might be attributed to the fact that: the target scene descriptors are very similar for instance of the same categories, while the predict scene descriptor is able to learn the pattern for one categories but not good at distinguishing the subtle difference among instances.}} The visualization of some results is shown in Figure \ref{fig:part_figure}.

\subsection{Ablation Study}\label{subsec:ablation}
To prove that scene descriptor could describe the global information and substitute the global category label, we conduct experiments on ShapeNet with SpiderCNN~\cite{xu2018spidercnn} as the backbone, {\color{black}since it} directly adopts category label. Then, without loss of generality, we conduct more ablation study on semantic segmentation task on ScanNet and choose PointConv as the baseline. 
\paragraph{SceneEncoder module versus Category information.}
For part segmentation{\color{black}, category labels are not always available, as for scene semantic segmentation, it is even hard} to give a label to a scene. By contrast, our SceneEncoder module learns to represent the global information with the scene descriptor extracted from the point cloud{\color{black}, even with no extra information of category label. To show the effectiveness of scene descriptor in representing global information,} we conduct the ablation study for different combinations with SpiderCNN as backbone. \cite{xu2018spidercnn} releases the result of SpiderCNN with category information, {\color{black}and we} conduct three extra experiments under the same setting, {\color{black}{\it i.e.,}} SpiderCNN with no category information, SpiderCNN with SceneEncoder module, and SpiderCNN with both category information and SceneEncoder module. As shown in Table \ref{tab:spidercnn}, \gjy{SpiderCNN with }SceneEncoder module \gjy{but }without category information perform as well as SpiderCNN with only category information. Therefore, {\color{black} it indicates} that the scene descriptor could also represent the global information and substitute the global category label. Additionally, compared with \cite{xu2018spidercnn}, combining the SceneEncoder module and category label can increase the class average IoU by $0.7\%$. This result {\color{black}illustrates} that the SceneEncoder could enhance the effect of global information.

\paragraph{SceneEncoder module and region similarity loss.}
\begin{table}
\centering
\begin{tabular}{lc} 
\toprule
Method  & mIoU \\
\midrule
PointConv & 55.6\\
\midrule
PointConv + SceneEncoder & 58.6 \\
PointConv + RSL & 58.7 \\
\midrule
PointConv + SceneEncoder + RSL & 62.8 \\
\bottomrule
\end{tabular}
\vspace{-2mm}
\caption{Ablation study for {\color{black}impact} of SceneEncoder module and region similarity loss.}
\vspace{-2mm}
\label{tab:pointconv}
\end{table}

To evaluate the {\color{black}impact} of the SceneEncoder module and the region similarity loss, we conduct two experiments {\color{black}on ScanNet}. First, we insert our SceneEncoder module into the PointConv and train the model without region similarity loss. Second, we just introduce the region similarity loss into the training process of the PointConv without the SceneEncoder module. The results are shown in Table \ref{tab:pointconv}, {\color{black}where we can observe that the SceneEncoder module and region similarity loss can individually improve the performance.} The combination can both deal with the global feature and local feature in a better way, hence {\color{black}further} improve the performance of PointConv by a large margin.

\paragraph{Selection {\color{black}s}trategy of region similarity loss.}
\begin{table}
\centering
\begin{tabular}{lc}  
\toprule
Selection Strategy  & mIoU \\
\midrule
Select points randomly & 60.2    \\
Select points with high confidence & 62.8\\
\bottomrule
\end{tabular}
\vspace{-2mm}
\caption{Ablation Study that whether choosing points randomly or according to the maximum classification confidence.}
\vspace{-4mm}
\label{tab:strategy}
\end{table}

In order to show the effectiveness of our selection strategy of distinguishing point features, we design experiments to study the performance of {\color{black}the region similarity loss} using another selection strategy. Namely, we select distinguishing point features among correctly classified points randomly. As shown in Table \ref{tab:strategy}, selecting distinguishing point features with high confidence could help to improve $2.5\%$ on mIoU, which proves the effectiveness of our proposed selection strategy.

\section{Conclusion}
In this paper, we propose a \gjy{scene-aware} semantic segmentation method with a SceneEncoder module and a region similarity loss. The SceneEncoder module makes full utilization of global information to predict a scene descriptor and this scene descriptor can aid the point level semantic segmentation by filtering out impossible results. To further refine the contour of segmentation, we propose the region similarity loss to {\color{black}propagate distinguishing features by forcing points in each local region with same labels to have similar final features.} Overall, the proposed methods achieve the state-of-the-art performance on ScanNet dataset for semantic segmentation and ShapeNet dataset for part segmentation. 
\bibliographystyle{named}
\bibliography{ijcai20-arxiv}

\end{document}